\documentclass[10pt,twocolumn,letterpaper]{article}

\usepackage{cvpr}
\usepackage{times}
\usepackage{epsfig}
\usepackage{graphicx}
\usepackage{amsmath}
\usepackage{amssymb}
\usepackage[pagebackref=true,breaklinks=true,letterpaper=true,colorlinks,bookmarks=false]{hyperref}
\usepackage{microtype}
\usepackage{booktabs}
\usepackage{tabularx}
\usepackage{soul}
\usepackage{url}

\graphicspath{{./figures/}}

\def\figurePath{figures/}
\def\myfigure#1#2{\begin{figure}[ht]\centering\includegraphics*[width = \linewidth]{\figurePath#1}\caption{#2}\label{fig:#1}\end{figure}}
\def\mycfigure#1#2{\begin{figure*}[tb]\centering\includegraphics*[clip, width = \linewidth]{\figurePath#1}\caption{#2}\label{fig:#1}\end{figure*}}

\def\mysection#1#2{\section{#1}\label{sec:#2}}

\def\mysubsection#1#2{\subsection{#1}\label{sec:#2}}
\def\mysubsubsection#1#2{\subsubsection{#1}\label{sec:#2}}

\newcommand{\refSec}[1]{Sec.~\ref{sec:#1}}
\newcommand{\refFig}[1]{Fig.~\ref{fig:#1}}
\newcommand{\refEq}[1]{Eq.~\ref{eq:#1}}
\newcommand{\refTbl}[1]{Tbl.~\ref{tbl:#1}}

\makeatletter
\renewcommand{\paragraph}{%
  \@startsection{paragraph}{4}%
  {\z@}{2.25ex \@plus 1ex \@minus .2ex}{-1em}%
  {\normalfont\normalsize\bfseries}%
}
\makeatother

\cvprfinalcopy

\def\cvprPaperID{211}

\ifcvprfinal\fi

\begin{document}

\title{Deep Reflectance Maps}

\author{Konstantinos Rematas\\
KU Leuven\\
{\tt\small krematas@esat.kuleuven.be}
\and
Tobias Ritschel\\
University College London\\
{\tt\small t.ritschel@ucl.ac.uk}
\and
Mario Fritz\\
MPI Informatics\\
{\tt\small mfritz@mpi-inf.mpg.de}
\and
Efstratios Gavves\\
University of Amsterdam\\
{\tt\small  E.Gavves@uva.nl}
\and
Tinne Tuytelaars\\
KU Leuven\\
{\tt\small Tinne.Tuytelaars@esat.kuleuven.be}
}

\maketitle

\begin{abstract}
Undoing the image formation process and therefore decomposing appearance into its intrinsic properties is a challenging task due to the under-constraint nature of this inverse problem.
While significant progress has been made on inferring shape, materials and illumination from images only, progress in an unconstrained setting is still limited.
We propose a convolutional neural architecture to estimate \emph{reflectance maps} of specular materials in natural lighting conditions.
We achieve this in an end-to-end learning formulation that \emph{directly} predicts a reflectance map from the image itself.
We show how to improve estimates by facilitating additional supervision in an \emph{indirect} scheme that first predicts surface orientation and afterwards predicts the reflectance map by a learning-based sparse data interpolation.

In order to analyze performance on this difficult task, we propose a new challenge of Specular MAterials on SHapes with complex IllumiNation (SMASHINg) using both synthetic and real images.
Furthermore, we show the application of our method to a range of image-based editing tasks on real images.

\end{abstract}

\section{Introduction}
A classic computer vision task is the decomposition of an image into its intrinsic shape, material and illumination.
The physics of image formation are well-understood: the light hits a scene surface with specific orientation and material properties and is reflected to the camera.\footnote{Project: \url{http://homes.esat.kuleuven.be/~krematas/DRM/}}
Factoring an image into its intrinsic properties, however, is very difficult, as the same visual result might be due to many different combinations of intrinsic object properties.

For the estimation of those properties, a common practice is to assume one or more properties as known or simplified and try to estimate the others.
For example, traditional approaches to intrinsic images or shape-from-shading assume lambertian materials, or point lights.
Furthermore, to simplify the problem, shape is often either assumed to be known in the form of a 3D model, or it is restricted to simple geometry such as spheres.

\begin{figure}[t]
\centering
\includegraphics[width=\linewidth]{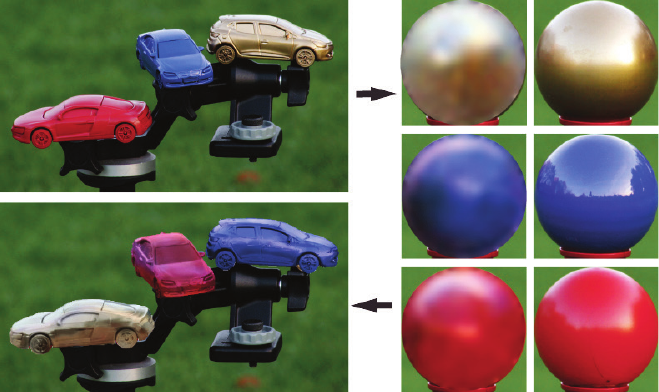}
\caption{\emph{Top:} Input 2D image with three cars of unknown shape and material under unknown natural illumination. \emph{Right:} Our automatically extracted reflectance map and the reference. \emph{Bottom:} Transfer of reflectance maps between the objects.}\label{fig:Teaser}
\end{figure}

In this work, we extract reflectance maps \cite{Horn1979} from images of objects with complex shapesand specular material, under complex natural illumination.
A reflectance map holds the orientation-dependent appearance of a fixed material under a fixed illumination.
It does not attempt to factor out material and/or illuminant and should not  be confused with a reflection map that contains illumination \cite{Debevec1998} or with surface reflectance \cite{Dana1999}.

Under the assumptions of a constant material, no shadows, a distant light source and a distant viewer, the relation of surface orientation and appearance is fully described by the reflectance map.
It can represent all illuminants and all materials, in particular specular materials under high-frequency natural illumination.
Therefore, besides allowing for a better understanding and analysis of 2D imagery, the ability to estimate reflectance maps lends itself to a broad spectrum of applications, including material transfer, inpainting, augmented reality and a range of image-based editing methods.

The input of our system is a 2D image where an object from a known class, (\eg, cars), was segmented (Fig~\ref{fig:Teaser}) and output is a reflectance map.
To this end, we propose two different approaches:
The first approach directly estimates a reflectance map from the input image using an end-to-end learning framework based on CNNs and deconvolutions.
The second approach decomposes the process in two steps, enabling the use of additional supervision in form of object surface normals at training time.
For the second approach we first predict per-pixel surface normals, which we use to compute sparse reflectance maps from the visible normals of the objects.
Given the sparse reflectance map, we introduce a learned sparse data-interpolation scheme in order to arrive at the final reflectance map.
In summary, we make the following five key contributions:

\begin{itemize}
\item First end-to-end approach to infer reflectance maps from a 2D image of complex shapes of specular materials under natural illumination.
\vspace{-0.15cm}
\item First dataset based on synthetic images and real photographs that facilitates the study of this task.
\vspace{-0.15cm}
\item
The first CNNs/deconvolutional architecture to  learn the complex mapping from the spatial 2D image to the spherical domain.
\vspace{-0.15cm}
\item The first CNN addressing a data-interpolation task of sparse unstructured data.
\vspace{-0.15cm}
\item Demonstration of our approach on a range of real images and a range of image-based editing tasks.

\end{itemize}

\mycfigure{Overview}{Overview of our approach, that comprises two variants: A direct one and an indirect one extracting surface orientations.}

\section{Related Work}
Factoring images into their constituting components is an important goal of  computer vision.
It is inherently hard, as many combinations of factors can result in the same image.
Having a decomposition available would help solving several important computer graphics and computer vision problems.

\paragraph{Factoring Images.}
Classic \emph{intrinsic images} factor an image into illuminant and reflectance \cite{Barrow1978}.
Similarly, \emph{shape-from-shading} decomposes into reflectance and shading, eventually leading to an orientation map or even a full 3D shape.
Larger-scale acquisition of reflectance \cite{Dana1999} and illumination \cite{Debevec1998} have allowed to compute their statistics \cite{Dror2001} helping to better solve inverse and synthesis problems.

Recently, factoring images has received renewed interest.
Lombardi and Nishino~\cite{Lombardi2015} as well as Johnson and Adelson~\cite{Johnson2011} have  studied the relation of shape, reflectance and natural illumination.
A key idea in their work is, that under natural illumination, appearance and orientation are in a much more specific relation (as used in Photometric stereo \cite{Hertzmann2005}) than for a single point light, where many similar appearance for totally different orientations can be present.
They present different optimization approaches that allow for high-quality estimation of one component if at least one other component is known.
In this work, we assume that the object is made of a single material and its object class and its segmentation mask are known.
However, we do not aim at factoring out illuminant, reflectance and shape, but keep the combination of reflectance and illuminant and only factor it from the shape.
Further factoring the reflectance map produced in our approach into material and illuminant would be complemented by methods such as \cite{Lombardi2015} or \cite{Johnson2011}.

The work of Baron and Malik~\cite{Barron2015a} factors shaded images into shape, reflectance and lighting, but  only for scalar reflectance, \ie diffuse albedo and for limited illumination frequencies.
In a very different vein, Internet photo collections of diffuse objects can be used to produce a rough 3D shape that serves extracting reflectance maps in a second step \cite{Haber2009}.

A recent approach by Richter and Roth~\cite{Richter2015} first estimates a diffuse reflectance map using approximate normals and then refines the normal map using the reflectance map as a guide.
Different from our approach, they assume diffuse surfaces to be approximated using 2nd-order spherical harmonics (SH) and learn to refine the normals from the reflectance map using a regression forest.
We compare the reflectance maps produced by our approach to reflectance maps using an SH basis which are limited to diffuse materials.

\paragraph{Computer Graphics.}
While appearance is considered view-independent in intrinsic images, view-dependent shading is described by \emph{reflectance maps} \cite{Horn1979}.
In computer graphics, reflectance maps are popular and known as  \emph{lit spheres} \cite{Sloan2001} or \emph{MatCaps} \cite{Zbrush}.
They are used to capture, transfer and manipulate the orientation-dependent appearance of photorealistic or artistic shading.
A special user interface is required, to map surface orientation to appearance at sparse points in an image, from which orientations are interpolated for in-between pixels to fill the lit sphere (\eg~\cite{RematasCVPR14} manually aligned a 3D model with an image to generate lit spheres).
Small diffuse objects in a single cluttered image were made to appear specular or transparent using image manipulations with manual intervention \cite{Khan2006}.
Our approach shares the simple and effective lit half-sphere parametrization but automates the task of matching orientation and appearance.

\paragraph{Deep Learning.}
In recent years convolutional neural networks (CNNs) have shown strong performance across different domains.
In particular, the strong models for object recognition~\cite{Krizhevsky2012} and detection~\cite{Girshick2014} can be seen as a layer-wise encoder of successively improved features.
Based on ideas of encoding-decoding strategies similar to auto-encoders, convolutional decoders have been developed \cite{zeiler2010deconvolutional,lee2009convolutional} to decode condensed representations back to images.
This has led to fully convolutional or deconvolutional techniques that have seen wide applicability for tasks where there is a per-pixel prediction target. In \cite{Long2015,hypercolumn}, this paradigm has been applied to semantic image segmentation.
In~\cite{Dosovitskiy2015} image synthesis was proposed given object class, view and view transformations as input and synthesizing segmented new object instances as output.
Similarly, \cite{Kulkarni2014} propose the \emph{deep convolutional inverse graphics networks} with an encoder-decoder architecture, that given an image can synthesize novel views.
In contrast, our approach achieves a new mapping to an intrinsic property -- the reflectance map.

\emph{Deep lambertian networks} \cite{Tang2012} apply deep belief networks to the joint estimation of a reflectance, an orientation map and the direction of a single point light source.
They rely on Gaussian Restricted Boltzmann Machines to model the prior of the albedo and the surface normals for inference from a single image.
In contrast, we address specular materials under general illumination, but without factoring material and illuminant.

Another branch of research proposes to use neural networks for
depth estimation~\cite{Eigen2014a,Li2015, Liu2015}, normal estimation~\cite{Eigen2015, Wang2015, Li2015}, intrinsic image decomposition\cite{Narihira2015b,Zhou2015} and lightness\cite{Narihira2015a}.
Wang~\etal~\cite{Wang2015} show that a careful mixture of deep architectures with hand-engineered models allow for accurate surface normal estimation.
Observing that normals, depth and segmentations are related tasks, \cite{Eigen2015} propose a coarse-to-fine, multi-scale and multi-purpose deep network that jointly optimizes depth and normal estimation and semantic segmentation.
Likewise, ~\cite{Li2015} apply deep regression using convolutional neural networks for depth and normal estimation, whose output is further refined by a conditional random field.
Going one step further, ~\cite{Liu2015} propose to embed both the unary and the pairwise potentials of a conditional random field in a unified deep network. In contrast, our goal is not normal, but rather reflectance map estimation. In particular, our ``direct approach'' makes do without any supervision of normal information, while the ``indirect approach'' has normals as a by-product. In addition, our new challenge dataset captures reflectance maps and normals for the specular case, which are not well represented in prior recordings -- in particular as also range sensors have difficulties on specular surfaces.

\section{Model}
\paragraph{Motivation}
We address a challenging inverse problem that is highly underconstrained. Therefore, any solution needs to mediate between evidence from the data and prior expectations -- in our case over reflectance maps. In the general settings of specular materials and natural illuminations, modeling prior expectations over reflectance maps -- let alone obtaining a parametric representation -- seems problematic. This motivated us to follow a data-drive approach in an end-to-end learning framework, where the dependence of reflectance maps on object appearances is learnt from a substantial number of synthesized images of a given object class.

\paragraph{Overview}

The goal of our network is the estimation of the reflectance map of an object depicted in a single RGB image (\refFig{Overview}).
This is equivalent to estimating how a sphere \cite{Sloan2001} with the same material as the object would look like from the same camera position and the same illumination.
From the estimated reflectance map, we can make the association between surface orientation and appearance.
This allows surface manipulation and transfer of materials and illumination between objects or even scenes.

We propose two approaches to estimate reflectance maps:  a direct (\refSec{DirectApproach}) and an indirect one (\refSec{IndirectApproach}).
Both have a general RGB image as input and a reflectance map as an output.
The indirect method also produces a conjoint per-pixel normal map.

Both variants are trained from and evaluated on the SMASHINg dataset introduced in detail in \refSec{Dataset}.
For now, we can assume the training data to consists of pairs of 2D  RGB images (domain) and reflection maps (range) in the parametrization explained in \refSec{Representation}.
This section now explains the two alternative approaches in detail.

\mysubsection{Reflectance Map Representation}{Representation}
A reflectance map  $L(\omega)\in\mathcal S^+\rightarrow\mathbb R^3$ \cite{Horn1979} is a map from orientations $\omega$ in the positive half-sphere $\mathcal S^+$ to the RGB radiance value $L$ leaving that surface to a distant viewer.
It combines the effect of illumination and material.
For the case of a mirror sphere it captures illumination \cite{Debevec1998} but is not limited to it.
It also does not capture surface reflectance \cite{Dana1999}, which would be independent of illumination, but joins the two.

There are multiple ways to parameterize orientation $\omega$.
Here Horn \cite{Horn1979} used positional gradients which are suitable for an analytic derivation but less attractive for computation as they are defined on the infinite real line.
We instead parameterize the orientation simply by $s,t$ the normalized surface normal's $x$ and $y$ components.
Dropping the $z$ coordinate is equivalent to drawing a sphere under orthographic projection with exactly this reflectance map as seen right in \refFig{Overview}.
Note, that orientations of surfaces in an image only cover the upper half-sphere, so we only need to parameterize a half-sphere, avoiding to deal with spherical functions, \eg spherical harmonics \cite{Richter2015}, that reduce the maximal frequency, only allowing for diffuse materials.

\mysubsection{Direct approach: End-to-end model for prediction of reflectance maps}{DirectApproach}
In the direct approach, we learn a mapping between the object image and its reflectance map, following a convolutional-deconvolutional architecture.

\myfigure{DirectArchitecture}{Architecture of the direct approach.}

The full architecture can be seen if figure~\refFig{DirectArchitecture}.
Starting from a series of convolutional layers, followed by batch normalization, ReLU and pooling layers, the size of the input feature maps is reduced to $1\times1$.
After continuing with two fully connected layers, the feature maps are  upsampled until the output size is $32\times32$ pixels.
In all convolutional layers a stride of 1 is used and padded with zeros such that the output has the same size as the input.
The final layer uses an euclidean loss between the RGB values for the predicted and the ground truth reflectance map.

In a typical CNN regression architecture, there is a spatial correspondence between input and output, \eg in normal or depth estimation or semantic segmentation.
In our case, the network needs to learn how to ``encode'' the input image so it can correspond to a specific reflectance map.
This task is particularly challenging as the model has to learn not only how to place the image pixels to locations in the sphere (change from image to directional domain), but also to impute and interpolate appearance for unobserved normals.

\mysubsection{Indirect approach: Reflectance maps from inferred normals and sparse interpolation}{IndirectApproach}
The indirect approach proceeds in four steps:
\emph{i)} estimating per-pixel orientation maps from the RGB image.
\emph{ii)} upsampling the orientation map to the full available input image resolution.
\emph{iii)} changing from the image domain into the directional domain, producing a sparse reflectance map.
\emph{iv)} predicting a dense reflectance map from the sparse one.

The first and fourth step are model by CNN architectures, while the second and third step are prescribed transformations, related to the parametrization of the reflectance map.
We will detail each step in the following paragraph.

\paragraph{Orientation estimation}
\myfigure{IndirectNormalArchitecture}{Architecture of the normal step of our indirect approach.}
Our goal in the first step is to predict a surface orientation map from the RGB image.
Thanks to our parametrization of the directional domain to coordinates in a flat 2D image of a lit sphere, the task is slightly simpler than finding full orientation.
We seek to find the $s$ and $t$ parameters according to our reflectance map parameterization.

We train a CNN to learn the $s,t$ coordinates.
The architecture of the network is shown in \refFig{IndirectNormalArchitecture}.
The network is fully convolutional as in \cite{Long2015} and it consist of a series of convolutional layers followed by ReLU and pooling layers that reduce the spatial extend of the feature maps.
After the fully convolutional layers, there is a series of deconvolutional layers that upscale the feature representation to half the original size.
Finally, we use two euclidean losses between the prediction and the $L_2$ normalized ground truth normals. The first one takes into account the $x,y,z$ coordinates of the normals, while the second only the $x,y$.

\paragraph{Orientation upsampling}
The orientations are estimated at a resolution of $n=128\times 128$, so the number of appearance samples is in the order of ten-thousands.
Most input images however are of much higher resolution with millions of pixels.
A full-resolution orientation map is useful for resolving all appearance details in the orientation domain. The appearance of one orientation in the reflectance map can be related to all high-resolution image pixels (millions).
Also intended applications performing shape manipulation in the 2D image (cf.~\refSec{QualitativeEvaluation}) will benefit from a refined map.
To produce a high-resolution orientation map, we used joint upsampling \cite{Kopf2007} as also done in range image \cite{Chan2008}.
Once we have an estimation of the object's normals, they can be mapped to a sphere and associated with appearance.

\paragraph{Change-of-domain}
We now reconstruct a sparse reflectance map from the orientation map and the input image.
This is a prescribed mapping transformation: The pairs of appearance $L_i$ and orientation $\omega_i$ in every pixel are unstructured samples of the continuous reflectance map function $L(\omega)$ we seek to recover. Our goal now is to map these samples from the image to the directional domain, constituting the reflectance map. The most straightforward solution is to perform scattered data interpolation, such as
\begin{align}
\label{eq:Reconstruction}
	L(\omega) =
	(
	\sum_{i=1}^n
	w(
	\left<
	\omega,
	\omega_i
	\right>)
	)^{-1}
	\sum_{i=1}^n
	w(
	\left<
	\omega,
	\omega_i
	\right>)
	L_i,
\end{align}
where $w(x) = \exp(-(\sigma \cos^{-1}(x))^2)$ is an RBF kernel.

In practice however, the orientation estimates are noisy and the requirements of a global reflectance map (infinite illumination, orthographic view, no shadows) are never fully met, asking for a more robust estimate.
We found darkening due to shadows to be the largest issue in practice.
Therefore, we perform a $\max$ operation over all samples closer than a threshold $\epsilon=\cos (5^\circ)$ instead of an average, as in
\[
	L(\omega) =
	\max
	\{
	w(
	\left<
	\omega,
	\omega_i
	\right>)
	L_i
	\},
	\quad
	w(x) =
	\begin{cases}
	1\quad\text{if } x > \epsilon\\
	0\quad\text{otherwise.}
	\end{cases}
\]
If one orientation is observed under different amounts of shadow, only the one that is not in shadow will contribute -- which is the intended effect. Still, the map resulting from this step is sparse due to normals that were not observed in the image as seen in \refFig{IndirectReflectanceMapArchitecture} (left).
This requires imputing and interpolating the sparse data in order to arrive at a dense estimate.

\paragraph{SparseNet: (Sparse-to-dense) Learning-based approach to sparse data interpolation}
The result of the previous step is a sparse reflectance map.
It is noisy due to errors from incorrect normal estimation and has missing information at orientations that were not observed in the image.
Note, that the latter is not a limitation of the normal estimation, but even occurs for ground truth surface orientations: If an orientation is not present, its appearance remains unknown.

A simple solution is to use \refEq{Reconstruction} which already provides a dense output.
We propose a learning-based approach to predict a dense reflectance maps from a sparse and noisy one.
Accordingly, the network is trained on pairs of sparse and dense reflectance maps.
The sparse ones are created using the first three steps explained (orientation from CNN, upsampling, Change-of-domain) on synthetic data where the target reflectance map is known by rendering a sphere.

\myfigure{IndirectReflectanceMapArchitecture}{Architecture of the reflectance map step of our indirect approach.}
The employed CNN architecture is shown in \refFig{IndirectReflectanceMapArchitecture}.
Input is the sparse reflectance map and output the dense one. We use the output of the convolutional layers as additional cue. After each deconvolution layer, we concatenate its output with the feature map from the respective convolution layer.
Again an $L_2$ loss between the output and the dense reference reflectance map is used.

\mysection{The SMASHINg Challenge}{Challenge}
We propose the Specular MAterials on SHapes with complex IllumiNation (SMASHINg) challenge. It includes
a dataset (\refSec{Dataset}) of real as well as synthetic images, groundtruth reflectance maps and normals (where available),
results from different methods to reconstruct (\refSec{Methods}) and
a set of metrics (\refSec{Metrics}) that we propose to evaluate and compare performance. At the time of publication we will make the data, baselines, our methods as well as the performance metrics publicly available.

\mysubsection{Dataset}{Dataset}
Our dataset combines synthetic images (\refSec{SyntheticImages}), photographs (\refSec{Photographs}) and images from the web (\refSec{InternetImages}) of cars.
All images are segmented into foreground and background.

\mysubsubsection{Synthetic images}{SyntheticImages}
Synthetic images are produced with random
\emph{i)} views,
\emph{ii)} 3D shapes,
\emph{iii)} materials,
\emph{iv)} illumination and
\emph{v)} exposure (\refFig{Dataset}) .
The view is sampled from a random position around the object, looking at the center of the object with a FOV of $40^\circ$.
The 140 3D shapes come from the free 3D Warehouse repository, indexed by Shapenet~\cite{LiSu2015}.
For each sample the object orientation around the $y$ axis is randomized.
Illumination is provided by 40 free HDR environment maps.
The exposure is sampled over the ``key'' parameter of Reinhard et al.'s photographic tone mapper \cite{Reinhard2002} between 0.4 and 0.6.
For materials, the MERL BRDF database \cite{Matusik2003} containing 100 materials is used.
Overall 60\,k sample images from that space are generated. We define a training-test split so that no shape, material or illumination is shared between the training and test set.
\myfigure{Dataset}{Our dataset comprises synthetic images with random view, 3D shape, material, illumination and exposure.}

\mysubsubsection{Photographs}{Photographs}
As real test images, we have recorded photos of six toy cars that were completely painted with a single car lacquer, placed in four  different lighting conditions and photographed from five different views, resulting a total of 120 images.
Additionally, those real images were manually segmented from the background.
A qualitative evaluation, including examples of such real images is found in the next section \refSec{QualitativeEvaluation}.

\mysubsubsection{Internet Images}{InternetImages}
In order to provide an even more challenging test set, we collect an additional 32 car images from Internet search. Here we do not have access to groundtruth normals or reflectance maps, but the test provides a realistic test case for imaged-based editing methods. Again, we have manually segmented out the body of the car. This allows the study of single material normal and reflectance map prediction\footnote{For the Internet Images we used networks that were trained on synthetic data from segmented meshes to contain only the body.}.

\mysubsection{Methods}{Methods}
{
\centering
\includegraphics*[width = \linewidth]{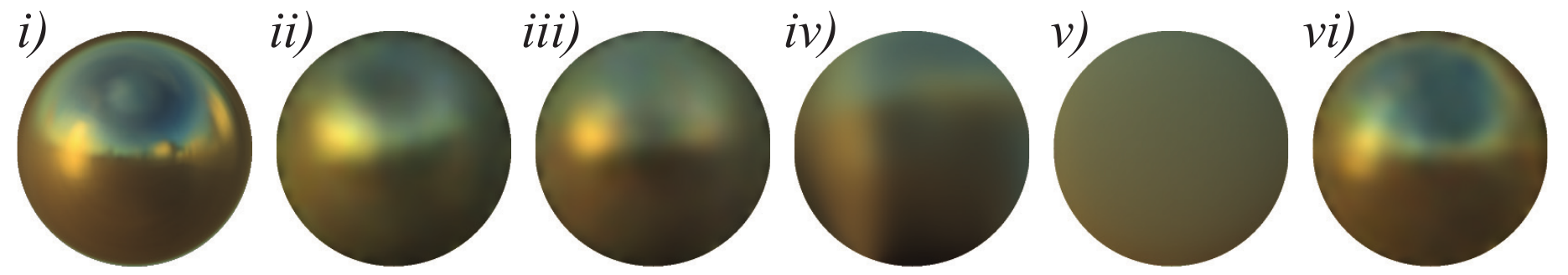}
}
We include six different methods to reconstruct reflectance maps:
\emph{i)} ground truth,
\emph{ii)} our direct ,
\emph{iii)} our indirect approach,
\emph{iv)} an approach that follows our indirect one, but does not use a CNN for sparse interpolation but an RBF reconstruction as described in \refEq{Reconstruction} (RBF),
\emph{v)} spherical harmonics (SH) where project the ground truth reflectance map to the SH domain,
\emph{vi)} an indirect approach where the estimated normals are replaced by ground truth normals.

\mysubsection{Metrics}{Metrics}
We employ two different metrics to assess the quality of reflectance map estimation. The first is plain $L_2$ error between all defined pixels of the reflectance map in RGB and the second the SSIM structural difference \cite{Wang2003}.

\mysection{Experiments}{Eperiments}
We evaluate our proposed end-to-end direct approach to reflectance maps on the new SMASHINg Challenge and compare it to the indirect approach in its different variants.
We start with a quantitative evaluation (\refSec{QuantitativeEvaluation}) followed by qualitative results in (\refSec{QualitativeEvaluation}) including a range of image-based editing tasks.

\mysubsection{Quantitative Results}{QuantitativeEvaluation}

\paragraph{Setup.}
Our quantitative results are summarized in \refTbl{ReflectanceError}. We provide results for our {\it Direct} method that learns to predict reflectance maps directly from the image in an end-to-end scheme, as well as several variants of our {\it Indirect} approach that utilizes intermediate result facilitated by supervision through normals at training time. The variants of the indirect scheme are based on our normal estimate, but differ in their second stage that has to perform a type of data interpolation to arrive at a dense reflectance map, given the sparse estimate.
For such interpolation scheme, we investigate the proposed learning-based approach {\it Indirect (SparseNet)} as well as using radial basis function interpolation {\it Indirect (RBF)}.
Furthermore, we provide best case analysis by using ground-truth normals in the indirect approach {\it Indirect (GT Normals)} (only possible for synthetic data) and computing a diffuse version of the ground-truth by means of spherical harmonics {\it GT (SH)}.
The latter gives an upper bound on the result that could be achieved by methods relying on a diffuse material assumption.

\begin{table}[h!]
	\small
	\centering
	\tabcolsep2pt
	\caption{Results for the different methods defined in \protect\refSec{Methods}.}
	\begin{tabular}{p{3.5cm}rrrr}
           & \multicolumn{2}{c}{Synthetic}& \multicolumn{2}{c}{Real} \\
           Method & MSE & DSSIM & MSE & DSSIM\\
	\toprule

	Direct      & .0019 	& .0209& .0120 & .0976\\
	Indirect (SparseNet)      & .0018 	& .0180 & .0143& .0991 \\
	Indirect (RBF) 	 & .0038 	& .0250 & .0116& .0814   \\
  \midrule
  	Indirect (GT Normals)  & .0008 	& .0111& ---& ---  \\
  	GT (SH)  & .0044 	& .0301 & .0114 & .0914  \\
	\bottomrule
	\end{tabular}
	\label{tbl:ReflectanceError}
\end{table}
\paragraph{Reflectance Map Analysis.}
Overall, we observe consistency among the two investigated metrics in how they rank approaches.
We obtain accurate estimations for the synthetic challenge set for our {\it direct} as well as the best {\it indirect} methods. The quantitative findings are underpinned by the visual results, e.g. showing the predicted reflectance maps in \refFig{MainResults}.
The performance on the real images is generally lower with the error roughly increasing by one order of magnitude. Yet, the reconstruction still preserve rich specular structures and give a truthful reconstruction of the represented material.

In more detail, we observe that the best {\it direct} and {\it indirect} approach perform similar on the synthetic data, although {\it direct} did not use the normal information during training.
For the real examples, this form of additional supervision seems to pay off more and even the simpler interpolations scheme RBF achieves best results in the considered metrics.
Closer inspection of the results clearly shows limitations of image-based metrics.
While the RBF-based technique yields a low error, it frequently fails to generate well localized highlight features on the reflectance map (see also illustration in \refSec{Methods}). We encourage the reader to visit the supplementary material, where a detailed visual comparison for all methods is provided.

The ground-truth baselines give further insights into improvements over prior diffuse material assumptions and the future potential of the method. The {\it GT (SH)} baseline shows that our best methods improve over a best-case diffuse estimate with a large margin on in the DSSIM metric -- highlighting the importance of considering more general reflectance maps. The error metric is again affected by the aforementioned issues.
The {\it Indirect (GT Normals)} illustrates a best case analysis of the indirect approach where we provide ground-truth normals.
The results show that there is potential to double the performance by having better normal estimation in the first stage.

\paragraph{Normal Analysis.}

\begin{table}[h!]
	\small
	\centering
	\caption{Normals estimation of indirect approach on synthetic data.}
	\begin{tabular}{llll}
          & Mean & Median & RMSE\\
	\toprule
	$L_2$ & 14.3 & 9.1 & 20.6\\
	Dual & 13.4 & 8.2 & 19.8\\
	Dual up & 13.3 & 8.2 & 19.9\\
	\bottomrule
	\end{tabular}
	\label{tbl:Normals}
\end{table}
\refTbl{Normals} quantifies the error in the normal estimate by the first stage of our indirect approach. This experiment is facilitated by the synthetic data where normals are available by the rendering pipeline.
$L_2$ corresponds to a network using the euclidean loss on the $x,y,z$ components of the normals, while dual uses the two losses described in \refSec{IndirectApproach}. Up refers to a network trained on upsampled normals.
Both, the dual loss and joint upsampling improve the estimation of normals.
Despite providing more data to the down-stream computation, the employed upsampling procedure does not decrease -- but rather slightly increase the accuracy of the normals.
While this analysis is conducted on synthetic data, we found that our models predict very convincing normal estimation even in the most challenging scenario that we consider, \eg \refFig{MainResults} and \refFig{ExchangeApplication}.

\mysubsection{Qualitative Results}{QualitativeEvaluation}
\mycfigure{MainResults}{
Results of different variants and steps of our approach \emph{(left to right)}.
Input image, GT RM, RM result of the direct approach, RM result of the indirect approach, reference RM, the sparse RM input produced in the indirect variant, and the normals produced by the indirect variant as well.
Each result is annotated to come from the synthetic, photographed or Internet part of our database.
For the Internet-based part, no reference RM is available.
Please see the supplemental material for exhaustive results in this form.
}

Automatically extracting reflectance maps -- together with the normal information we get as a by-product -- facilitate a range of image-based editing applications, such as material acquisition, material transfer and shape manipulation. In the following, we present several example applications. The supplementary material contains images and videos that complement our following presentation.

\paragraph{Reflectance Map and Normal Estimation.}
Typical results of estimated reflectance maps are presented in \refFig{MainResults}, also showing the quality of the predicted normals. The first row shows two examples on synthetic images, the second and third row show examples on real images and the last row shows examples of web images (no reference reflectance map is available here). Notice how the overall appearance, reflecting the interplay between material and the complex illumination, is captured by our estimates. In most examples, highlights are reproduced and even a schematic structure of the environment can be seen in the case of very specular materials.

\paragraph{Material Acquisition for Virtual Objects.}
\refFig{RealToVirtual} shows  synthesize image (column 2-5) that we have rendered from 3D models using the reflectance map automatically acquired from the images in column 1. Here, we use ambient occlusion \cite{Zhukov1998} to produce virtual shadows. This application shows how material representations can be acquired from real objects and transferred to a virtual object. Notice how the virtual objects match in material, specularity and illumination to the source image on the left.

\myfigure{RealToVirtual}{Transfer of reflectance maps from real photographs \emph{(1st col.)} to virtual objects \emph{(other col.'s)} of the same and other shape.
The supplemental video shows animations of those figures.}

\paragraph{Material Transfer.}
In order to transfer materials between objects of a scene, we estimate reflectance maps for each object independently, swapped the maps, and then use the estimated normals to re-render the objects using a normal lookup from the new map.
To preserve details such as shadows and textures, we first re-synthesize each object with its original reflectance map, save the per-pixel difference in LAB color space, re-synthesize with the swapped reflectance map and add back the difference in LAB.
An example is shown in \refFig{ExchangeApplication}.
Despite the uncontrolled conditions, we achieve photorealistic transfer of the materials -- making it hard to distinguish source from target.
\myfigure{ExchangeApplication}{
Material transfer application:
Images on the diagonal are the original input.
Off-diagonal images have the material of the input in its column combined with the input shape of its row.}

\paragraph{Shape Manipulation.}
As we estimate reflectance maps and surface normals, this enables various manipulation and re-synthesis approaches that work in the directional or normal domain.
Here, the surface orientation is changed, \eg using a painting interface and new appearance for the new orientation can be sampled from the reflectance map.
Again, we save and restore the delta of the original reflectance map value and the re-synthesized one to keep details and shadows.
An example is shown in \refFig{ShapeManipulationApplication}. The final result gives a strong sense of 3D structure while maintaining an overall consistent appearance w.r.t. material and scene illumination.
\myfigure{ShapeManipulationApplication}{
Shape manipulation application.
A user has drawn to manipulate the normal ma extracted from our indirect approach.
The reflectance map and the new normal map can be used to simulate the new shape's appearance.
Please see the supplemental video for a live demo of direct shape manipulation in the photo.}

\mysection{Discussion}{Discussion}
While our approach addresses a more general setting than previous methods, we still make certain assumptions and observe limitations:
\emph{i)} Up to now, all our experiments were conducted on cars and we assume that the car is segmented or in uniform background.
Yet, our approach is learning-based and should -- in principle -- adapt to other classes in particular given dedicated training data.
\emph{ii)} We assume that the object is made out of a single material and up to now we cannot handle multiple parts or textures.
\emph{iii)} We assume distant illumination and therefore light interaction of close by objects or support surfaces (\eg road) cannot be accurately handled by our model.
\emph{iv)} Due to our target representation of reflectance maps, the illumination is ``baked in'' and surface reflectance and illumination cannot be edited separately.
\emph{v)} Our quantitative evaluations are limited due to the absence of reliable (\eg perceptual) metrics of reflectance maps.

\mysection{Conclusion}{Conclusion}
We have presented an approach to estimate reflectance maps from images of complex shapes with specular materials under complex natural illumination.
We are not aware of previous attempts to solve this task.
Our study is facilitated by a new benchmark of synthetic, real and web images of increasing difficulty, that we will make available to the public. Our approach features the first mapping using end-to-end learning from image to the directional domain as well as an application of neural networks to learning-based sparse data interpolation. We show how to incorporate additional supervision by normal information that increase accuracy as well as results in normal estimations as a byproduct. Our results show truthful reflectance maps in all three investigated scenarios and we demonstrate the applicability on several image-based editing tasks.

{\small
\bibliographystyle{ieee}
\bibliography{DeepReflectanceMaps}
}

\end{document}